# Bubble2Heat: Optical to Thermal Inference in Pool Boiling Using Physics-encoded Generative AI

Qianxi Fu[1], Youngjoon Suh[1], Xiaojing Zhang[1], Sanghyeon Chang[1], Yoonjin Won[1,2,*]

[1]Department of Mechanical and Aerospace Engineering, University of California, Irvine, Irvine, CA 92697, USA
[2]Department of Electrical Engineering and Computer Science, University of California, Irvine, Irvine, CA 92697, USA

*Corresponding Authors Email: won@uci.edu

## Abstract

Phase change process plays a critical role in thermal management systems, yet quantitative characterization of multiphase heat transfer remains limited by the challenges of measuring temperature fields in chaotic, rapidly evolving flow regimes. While computational methods offer temperature data at a high spatiotemporal resolution in ideal cases, replicating complex experimental conditions remains prohibitively difficult. In this paper, we present a deep learning framework that can generate temperature field data at simulation resolution from segmented high-speed recordings and pointwise thermocouple readings which are typically available in a canonical pool boiling experimental configuration without requiring advanced techniques. This framework leverages a conditional generative adversarial network trained only on simulation data. To ensure direct applicability of the model to experimental data, our framework also introduces a preprocessing pipeline that aligns high resolution simulation data with experimental measurements through both conventional image processing and image segmentation with pretrained convolutional neural network. We further show that standard data augmentation strategies are effective in enhancing the physical plausibility of the inference when precise physical constraints are not applicable. Our results highlight the potential of deep generative models to bridge the gap between observable multiphase phenomena and underlying thermal transport, offering a powerful approach to augment and interpret experimental measurements in complex two-phase systems.

# 1. Introduction

Phase change process plays a vital role in thermal management systems across a wide spectrum of technologies from industrial power generation [1] to electronics cooling [2]. Gaining a detailed understanding of the underlying thermal transport processes during phase change is essential for optimizing performance and ensuring reliability. A central challenge in advancing the science of boiling lies in the difficulty of capturing key thermal fields, particularly temperature, with sufficient spatial and temporal resolution. Although engineering design has traditionally relied on empirical correlations and theoretical models informed by experimental observations, these approaches depend on measurable quantities that remain elusive in complex, multiphase environments. Unlike single-phase systems, multiphase systems such as pool boiling involves rapid phase transitions, interfacial instabilities, and transient bubble dynamics that render conventional measurements difficult to obtain [3]. Consequently, the absence of high-resolution thermal field data continues to hinder our ability to obtain deeper physical insight and the development of more predictive modeling frameworks.

To address the difficulty of capturing thermal field data in pool boiling, various experimental diagnostics have been developed to extract localized or indirect measurements of heat transfer. Microheater arrays with integrated heat flux sensors, for example, allow pointwise estimation of thermal transport near nucleation sites, offering insight into microlayer evaporation and surface convection [4]. High-speed imaging enables visualization of vapor dynamics and bubble interactions, which can help infer heat transfer modes [5]. Infrared thermography, when synchronized with visual imaging, has further enhanced our ability to track surface temperature variations over time [6]. More recent advanced techniques such as planar laser induced fluorescence (PLIF) have been developed to directly measure the temperature distribution within multiphase structures such as droplets [7], liquid films [8,9] and spray mixture [10]. While each of these methods has yielded valuable observations, they remain constrained by the complex and delicate experimental setup required to achieve high spatial and temporal resolution [3]. Most importantly, they do not provide dense, field-level thermal data in a form suitable for generalized modeling or data-driven learning. These limitations motivate the search for alternative approaches capable of extracting richer thermal information from accessible measurements.

Given the challenges of acquiring comprehensive thermal data experimentally, simulations have been used to gain insight into boiling heat transfer by resolving fluid and thermal fields under controlled conditions. Computational fluid dynamics (CFD), including methods such as volume-of-fluid, level set, and front-tracking [11], has enabled detailed studies from single-bubble growth to complex bubble interactions [12-14]. The lattice Boltzmann method (LBM) in particular has gained attention for its ability to simulate complex interfacial phenomena without explicit interface tracking [15,16]. Furthermore, the mesoscopic nature of LBM makes it a balanced CFD method both in terms of accuracy and computational speed. While these techniques have demonstrated qualitative and even quantitative agreement with certain experimental results, they remain limited in their broader applicability. Accurately capturing multiphase dynamics requires careful modeling of interfacial motion, latent heat transfer, and variable material properties, which are all highly sensitive to numerical resolution and boundary conditions. Moreover, boiling spans a wide range of spatial and temporal scales that are difficult to capture within a single computational framework. High-fidelity simulations often suffer from high computational cost, limited convergence, and difficulty in handling realistic geometry or material variability [10]. These challenges restrict the ability of traditional CFD methods to provide generalizable, high-throughput simulations for real-world boiling systems.

Recent deep learning technologies have emerged as a powerful tool for modeling complex physical systems that are difficult to describe analytically or simulate with high fidelity. Its generalizability and efficiency have delivered promising results as more research is devoted to the application of deep learning in a variety of multiphase systems [17]. Specifically, deep learning-assisted computer vision models have shown to be capable of automating data collection tasks such as bubble detection and phase segmentation [18-26], greatly reducing manual effort in analyzing visual datasets. This allows the access to additional structural and statistical data that is previously difficult to obtain. Moreover, it has enabled effective training of neural networks on large datasets to predict quantities such as heat transfer coefficients [21,23,26] and perform flow regime classification [18,20,23,24]. In parallel, physics-informed neural networks (PINNs) have shown promise as surrogate models by embedding governing equations into the learning process, offering improved physical consistency in the absence of dense data [27]. However, PINNs often require known boundary and initial conditions which may not be available, and can struggle with convergence or accuracy in chaotic, data-sparse multiphase systems [28,29].

In contrast, generative deep learning models such as generative adversarial networks (GANs) [30] excel at learning mappings directly from data that can possess complex features, without requiring full explicit knowledge of the underlying physics and patterns. This property of generative models also provides high customizability in types of output through additional explicit input labels, such as in the case of conditional GAN (CGAN) [31], which have shown to be effective in tasks such as image-to-image translation [32] and high-resolution image generation [33,34]. In the realm of heat transfer and fluid dynamics, GANs including their conditional variants have been readily employed in generating high-resolution 3D chaotic turbulent flow [35] and temperature data inference in single-phase system [36]. Furthermore, recent work such as inference on electromagnetic properties in three-phase flow [37] and flow rate prediction in gas-liquid mixture [38] has motivated investigation of the applicability of generative neural networks in multiphase systems.

In this work, we present a deep learning framework for inferring 2D temperature fields in pool boiling systems using only easily measurable data including visual information of the phase geometry and pointwise temperature measurements. We propose and analyze a physics-encoding CGAN (PECGAN) model to learn and represent the mapping between multiphase geometry and the underlying temperature distribution, where conditional labels in the form of additional temporal geometric information and temperature references specify the desired outputs. To train the model, we generate paired temperature and phase contour data, which serves as the only viable ground truth for supervised learning, using a hybrid lattice Boltzmann and finite difference (LB-FD) method [39] that capitalizes on the inherent efficiency of the simulation method. Once trained, the model is applied directly to experimental data where the encoded physics is transferred, enabling temperature field inference without the need for intrusive and complex measurements. To enhance the physical consistency and improve the robustness of the inference result, we further apply data augmentation techniques as a substitute to hard physical constraints that are not available for our use case. This approach represents a step forward in bridging simulation-based learning with experimental analysis, allowing us to infer otherwise inaccessible thermal information from readily obtainable visual data. This paper details the model architecture, training process, and evaluation results, and explores the broader implications of applying deep generative models to multiphase heat transfer problems.

## 2. Framework

We develop a deep learning framework to infer temperature fields in pool boiling systems using only high-speed visual data and minimal thermal measurements following the general approach outlined in Fig. 1. The framework consists of data generation, data processing, training, and testing via experimental data. (i) We first generate data via simulation from a LB-FD simulation method [39], which provides paired temperature and phase contour maps that serve as the ground truth. We in addition collect high-speed recordings and thermocouple readings at the heater of a pool boiling system at varying heat fluxes as our experimental data. (ii) We preprocess both simulation and experimental data into a shared input format consisting of a sequence of phase contour maps, denoted by $\phi$; and a reference temperature map based on the average excess temperature, denoted by $T_{\text{ref}}$, prior to training. For simulation data, we derive phase contour maps by thresholding the density field to separate solid, liquid, and vapor regions. We construct the reference temperature map using the associated spatiotemporally averaged excess temperature and saturation temperature. (iii) We perform model training, where simulation physics is implicitly encoded into the PECGAN (Fig. 1(a)); and model testing, where the trained network is applied to the experimental data to output inferences based on the encoded physics (Fig. 1(b)). To enable application to experimental observations, we process experimental data into the same input structure as the simulation data: we extract phase contours from high-speed images by instance segmentation using a pre-trained Mask R-CNN [23,40] and manually construct the reference temperature maps from thermocouple measurements. This consistent data formatting bridges experimental and simulation data, using the bubble masks to infer critical thermal-fluid properties in the surrounding regions, providing a novel pathway for analyzing bubble dynamics and their impact on the liquid environment.

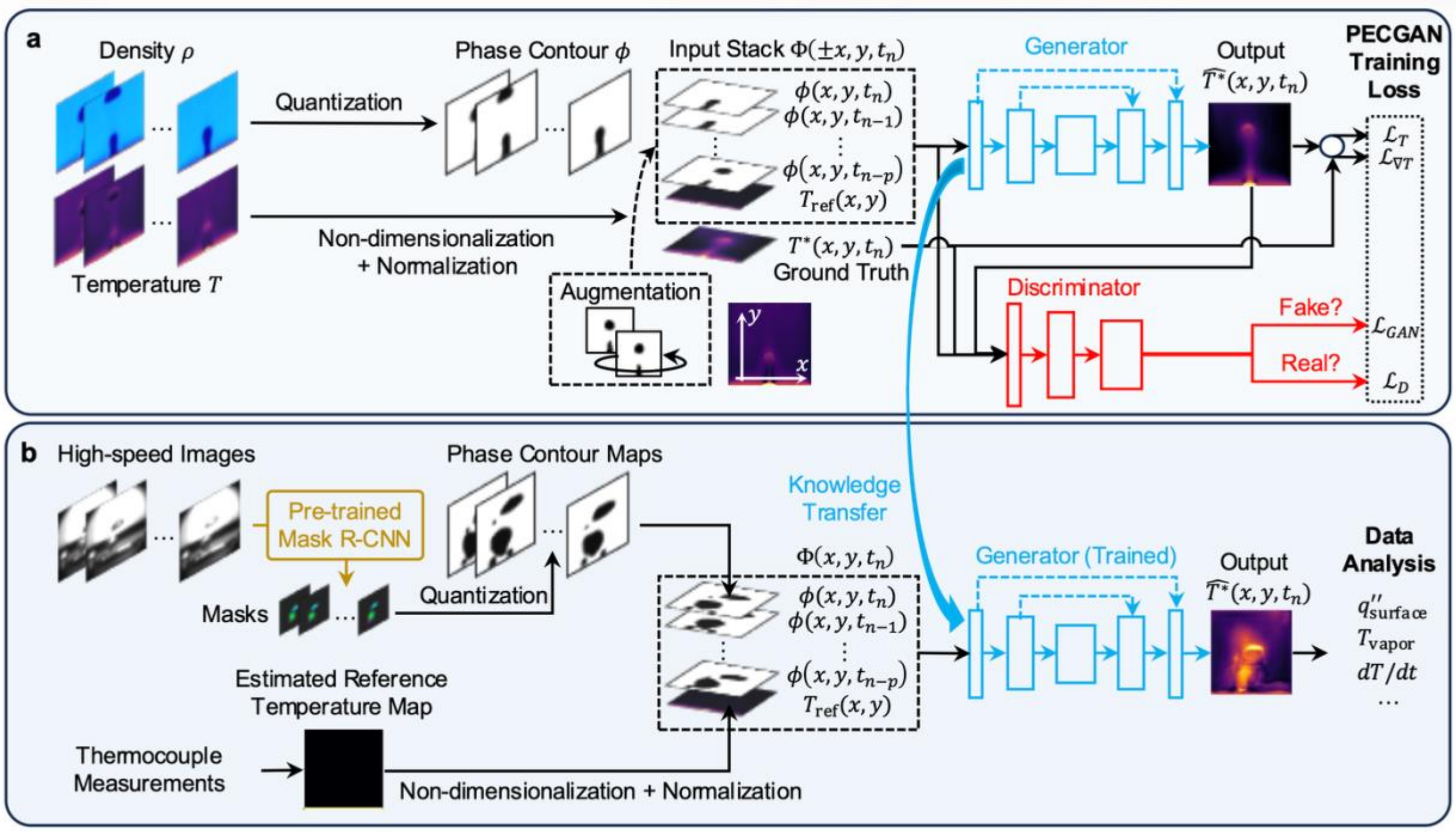

**Fig. 1.** The deep learning framework for temperature data generation. (a) The training stage where the physics-encoding conditional generative adversarial network (PECGAN) is trained on simulation data. The input stack is composed of the current phase contour, a sequence of past phase contours, and a reference temperature map $T_{\text{ref}}$ representing the saturation temperature and average excess temperature (see Methods). (b) The testing stage where temperature inferences are made by applying trained PECGAN to preprocessed experimental data, which enables further thermal analysis.

# 3. Methods

### 3.1 Simulation Dataset and Preprocessing

We simulate 2D pool boiling with a resolution of 256 x 256 lattice units (lu) using a hybrid LB-FD simulation. For simplicity, we adopt common simulation conditions (see Supplemental Information 1) which has shown to be stable [38]. Thermal simulation conditions include Dirichlet boundary conditions in the bottommost solid layer for temperature, and uniform initial bulk liquid temperature at a saturation temperature of $T_{\text{sat}} = 0.9T_c = 0.0656$ in lattice temperature unit (tu). A total of 9 sets of simulation data is generated with varying heater temperature and configurations for a diverse training dataset. Each set contains 200 frames. All heaters have a length of 40 lu and are arranged 40 lu apart from each other if there are multiple as shown in Fig. 2(a). Table 1 records the variation in heater count and heater temperature among different sets.

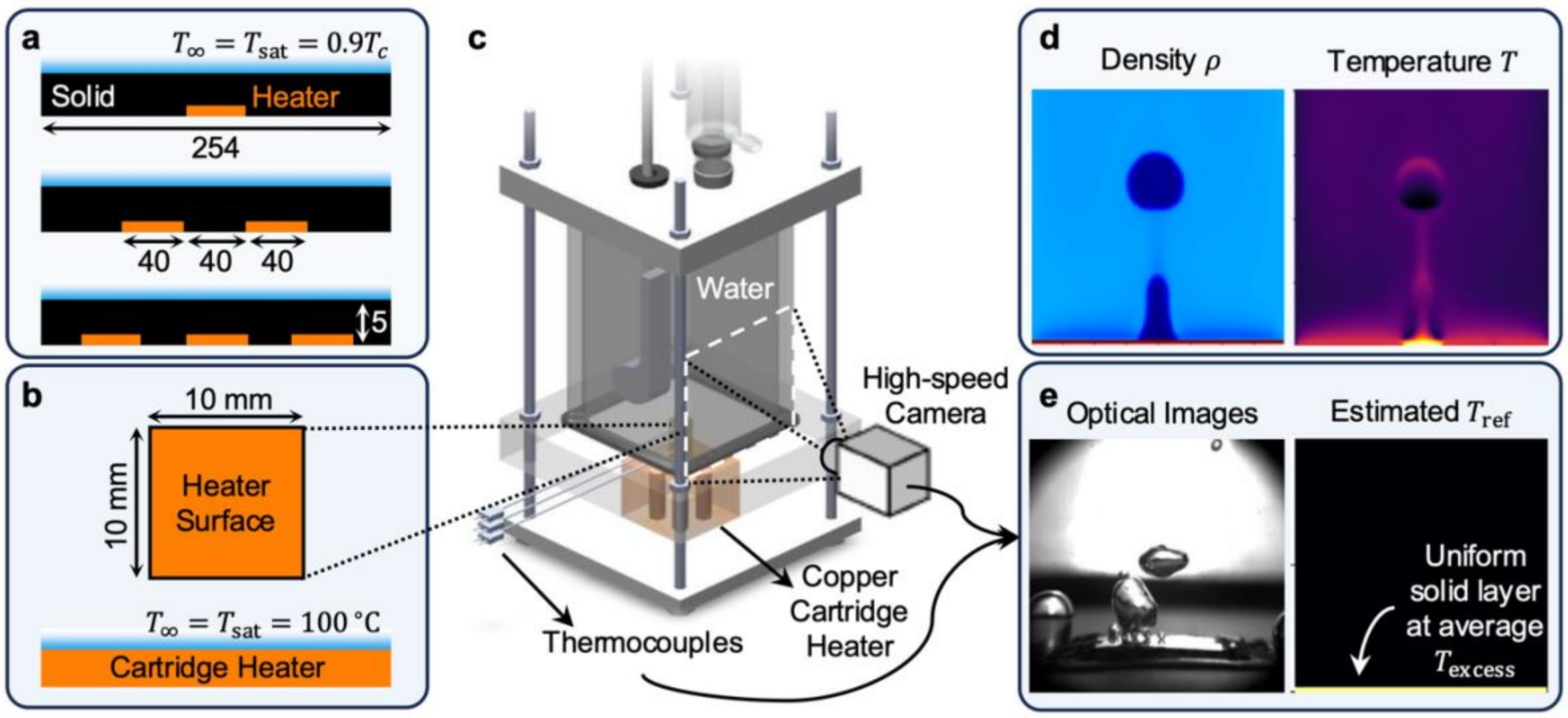

**Fig. 2.** Types of data collected for model training and testing. Heater configurations and ambient conditions used in (a) lattice Boltzmann and finite difference (LB-FD) simulation and (b) experiments. (c) Physical model of the experimental setup. (d) Types of LB-FD simulation data and (e) types of experimental data used in model training and testing.

To obtain phase contour maps $\phi$ from simulation, we quantize the density map $\rho$ using a global threshold $\rho_{\text{th}}$ derived from density distribution (see Supplemental Information 1), and manually assign values to each corresponding phase based on the following formula:

$$\phi(x, y, t) = \begin{cases} 0, & \text{if } \rho \leq \rho_{\text{th}} \text{ (vapor)} \\ 1, & \text{if } \rho > \rho_{\text{th}} \text{ (liquid)} \\ -1, & \text{for solid} \end{cases} \tag{1}$$

To normalized temperature values, we convert all temperature maps $T$ to a normalized (modified) Jakob number $T^*$ using the following equation:

$$T^* = \frac{c_{p,l}(T - T_{\text{sat}})}{h_{fg}} \frac{1}{\text{Ja}_{\text{max}}} = \frac{\text{Ja}}{\text{Ja}_{\text{max}}} \tag{2}$$

where $c_{p,l}$ is the heat capacity of liquid water, $h_{fg}$ is the latent heat of evaporation. All fluid properties are taken at saturation temperature, and the maximum Jakob number $\text{Ja}_{\text{max}}$ is taken to be 0.222 from Table 1.

**Table 1**. Heater conditions used in each simulation dataset.

| Dataset | Heater Count | $T_{\text{heater}}$ (tu) | $T_{\text{heater}}$ (K) | $\text{Ja}_{\text{heater}}$ | $T^*_{\text{heater}}$ |
|---|---|---|---|---|---|
| S1 | 1 | 0.074 | 656.94 | 0.150 | 0.676 |
| S2 | | 0.076 | 674.69 | 0.186 | 0.838 |
| S3 | | 0.078 | 692.45 | 0.222 | 1.000 |
| S4 | 2 | 0.074 | 656.94 | 0.150 | 0.676 |
| S5 | | 0.076 | 674.69 | 0.186 | 0.838 |
| S6 | | 0.078 | 692.45 | 0.222 | 1.000 |
| S7 | 3 | 0.074 | 656.94 | 0.150 | 0.676 |
| S8 | | 0.076 | 674.69 | 0.186 | 0.838 |
| S9 | | 0.078 | 692.45 | 0.222 | 1.000 |

### 3.2 Experimental Dataset and Preprocessing

We conduct pool boiling experiments and record 3 sets of high-speed images of the bubble phenomenon with average surface heat fluxes ranging from 5.24 W/cm$^2$ to 22.92 W/cm$^2$. Each set of high-speed image sequence consists of 500 images with a resolution of 1,024 x 1,024 pixels at 2,000 frames per second (fps). Thermocouple measurements are also taken within the copper heater cartridge to compute the corresponding heat flux to the system, as shown in Fig. 2(c).

To obtain the phase contour maps from high-speed images, we perform instance segmentation through a pre-trained Mask R-CNN model [23]. Outputs are in the form of instance-specific masks over bubbles, isolating vapor region from the surrounding liquid region. Values within each phase region are then assigned according to Eq. (1). Solid heater layer position is estimated from the original image and manually added after segmentation (see SI-2). We manually construct the initial temperature map by estimating the heater temperature through thermocouple measurements under the assumptions that the heater temperature remains uniform, and the bulk liquid temperature is initially at a saturation temperature $T_{\text{sat}} = 100\ ^\circ\text{C}$. The reference temperature map is then normalized through Eq. (2).

**Table 2**. Heater condition used in each experimental dataset.

| Dataset | Heater Dimension (mm × mm) | Heater Power (W) | Measured Heat Flux $q''$ (W/cm$^2$) | $T_{\text{heater}}$ (K) | $\text{Ja}_{\text{heater}}$ | $T^*_{\text{heater}}$ |
|---|---|---|---|---|---|---|
| E1 | 10 × 10 | 30 | 5.24 | 385.89 | 0.0238 | 0.107 |
| E2 | | 50 | 14.33 | 388.88 | 0.0294 | 0.133 |
| E3 | | 70 | 22.92 | 390.90 | 0.0332 | 0.150 |

### 3.3 Model Structure and Training

We define our PECGAN architecture using a U-Net-like generator [32,41] and a simple feedforward discriminator, as illustrated in Fig. 1. The generator learns an explicit mapping from input maps to the corresponding temperature field, while the discriminator implicitly learns the joint structure of the inputs and outputs by analyzing both phase contour and temperature maps together. This adversarial setup encourages the generator to produce temperature fields that are not only numerically accurate but also physically consistent with realistic phase dynamics. The training objective (see SI-3) of the generator $L_G$ is a weighted combination of the following loss components: temperature loss $L_T$ and gradient loss $L_{\nabla T}$, which minimizes pixel-wise differences between predicted and ground truth temperature maps and associated image gradients (here loss in horizontal gradient and vertical gradient are computed separately instead of computing loss in magnitudes. See SI-3), and an adversarial loss $L_{GAN}$, which represents the dissimilarity between the generated images and the real images as recognized by the discriminator:

$$L_G = \lambda_T L_T + \lambda_{\nabla T} L_{\nabla T} + L_{GAN} \tag{3}$$

where $\lambda_T = 10$ and $\lambda_{\nabla T} = 1$ are the weights assigned to the respective loss components. Here, we employ mean square error for both reconstruction loss and boundary loss, and binary cross-entropy for the adversarial loss. The training objective of the discriminator is to maximize the adversarial loss and conditional loss $L_C$ which represents its inability to label real images as real:

$$L_D = -L_{GAN} + L_C \tag{4}$$

To improve model robustness and generalization, as well as to compensate for the lack of imposable physical constraints, we investigate the effect of enforcing perspective invariance. This is implemented as a data augmentation technique, specifically by horizontally mirroring the input contour maps during training. We denote models with data augmentation applied using a superscript + sign. This strategy is compatible with the periodic boundary conditions of our simulation datasets. We aim to encourage the model to produce consistent inference regardless of viewing orientation.

To evaluate the model's ability to generalize across varying heater conditions, we train on simulation datasets S1, S3, S4, and S6, which collectively span the full range of heater temperatures and include two distinct heater geometries. As shown in Fig. 1, each input stack includes a set of phase contour maps and a normalized initial temperature map. For the results presented in this work, we vary the input map count, denoted by $p$, to provide temporal context.

# 4. Results and Discussion

### 4.1 Model Performance Evaluation with Simulation Data

We first evaluate the performance of trained generators on all simulation datasets. We report the mean absolute error of temperature, magnitude of the gradient of temperature, and the successive difference in temperature across time frames in Table 3. We note that the inclusion of additional phase contour maps (i.e., increasing $p$) is effective in reducing the inference error in most error metrics. Specifically, generator $G_{p=3}$ achieves lower error in all metrics but two compared to generators with lower $p$ values from Table 3. Figs. 3(a)-(d) also show a general decrease in error and error variance with an increase in $p$. The qualitative effect of increasing $p$ can be observed in Fig. 3(g). For simulation datasets, applying quantization may result in unstable bubble contours (i.e., the vapor phase contour diminishes or disappears entirely as bubble propagates). Consequently, the model is unable to produce temperature values that corresponds to vapor region since the model does not observe any vapor phase contours in the case of one-to-one image-to-image translation. With larger $p$ however, the model has access to a history of bubble trajectory, which enables inference even if bubble contour eventually degenerates. We observe that generators with larger $p$ tend to have the lowest temporal error among all generators as a result of using a sequence of phase contours rather than singular image as input. The disadvantage of increasing $p$ is that the error and error variance in the inferred temperature gradient will increase as shown in Figs. 3(e) and 3(f). This is because having more input information induces more noise locally within the generated temperature profile, which affects the predicted gradient. Adding a reference temperature map as an input is shown to slightly stabilize the error in gradient as shown in Figs. 3(e) and 3(f) without any significant change to other inference error metrics.

**Table 3**. Mean absolute error of the inferred $T^*$ across different models. Modifications to models are notated using subscripts $p$, which denotes the number of past phase contours included in the input stack; and $T_{\text{ref}}$, which denotes the inclusion of a reference temperature map. The superscript + denotes applied data augmentation during model training. Values are averaged across all 9 simulation datasets (see SI-4 for error values in training datasets and testing datasets separately). Fluid region is the combined region of liquid and vapor. The smallest and the second/third smallest value in each column are bolded and underlined, respectively. Note that unity in $T^*$ corresponds to a temperature of 0.0124 tu from Eq. (2), or 110.19 K by the principle of corresponding states [42].

| Model | **Average $\varepsilon_T$** | | | **Average $\varepsilon_{\|\|\nabla T\|\|}$ (pixel$^{-1}$)** | | | **Average $\varepsilon_{\partial T/\partial t}$ (frame$^{-1}$)** | |
|---|---|---|---|---|---|---|---|---|
| | Solid | Liquid | Vapor | Solid | Surface | Fluid | Solid | Fluid |
| $G_{p=0}$ | 0.0594 | 0.0133 | 0.0562 | <u>0.0112</u> | <u>0.00869</u> | <u>0.00107</u> | 0.00937 | 0.00366 |
| $G_{p=1}$ | 0.0578 | 0.0124 | 0.0534 | 0.0122 | **0.00805** | 0.00114 | 0.00770 | 0.00331 |
| $G_{p=2}$ | **0.0541** | 0.0135 | 0.0504 | 0.0131 | <u>0.00919</u> | 0.00117 | 0.00656 | 0.00281 |
| $G_{p=3}$ | <u>0.0556</u> | <u>0.0117</u> | <u>0.0424</u> | **0.0102** | 0.00982 | 0.00109 | 0.00536 | **0.00256** |
| $G_{p=2,T_{\text{ref}}}$ | 0.0572 | **0.0109** | 0.0501 | 0.0114 | 0.00943 | **0.00102** | 0.00632 | <u>0.00262</u> |
| $G_{p=3,T_{\text{ref}}}$ | <u>0.0552</u> | 0.0118 | <u>0.0432</u> | 0.0114 | 0.00932 | <u>0.00107</u> | 0.00570 | <u>0.00262</u> |
| $G^{+}_{p=2}$ | 0.0581 | 0.0138 | 0.0482 | 0.0135 | 0.01024 | 0.00116 | 0.00612 | 0.00301 |
| $G^{+}_{p=3}$ | 0.0565 | 0.0122 | 0.0458 | 0.0125 | 0.01169 | 0.00120 | <u>0.00500</u> | 0.00269 |
| $G^{+}_{p=2,T_{\text{ref}}}$ | 0.0597 | 0.0125 | 0.0469 | <u>0.0105</u> | 0.01215 | 0.00115 | **0.00483** | 0.00267 |
| $G^{+}_{p=3,T_{\text{ref}}}$ | 0.0568 | <u>0.0117</u> | **0.0395** | 0.0134 | 0.01427 | 0.00121 | <u>0.00514</u> | 0.00263 |

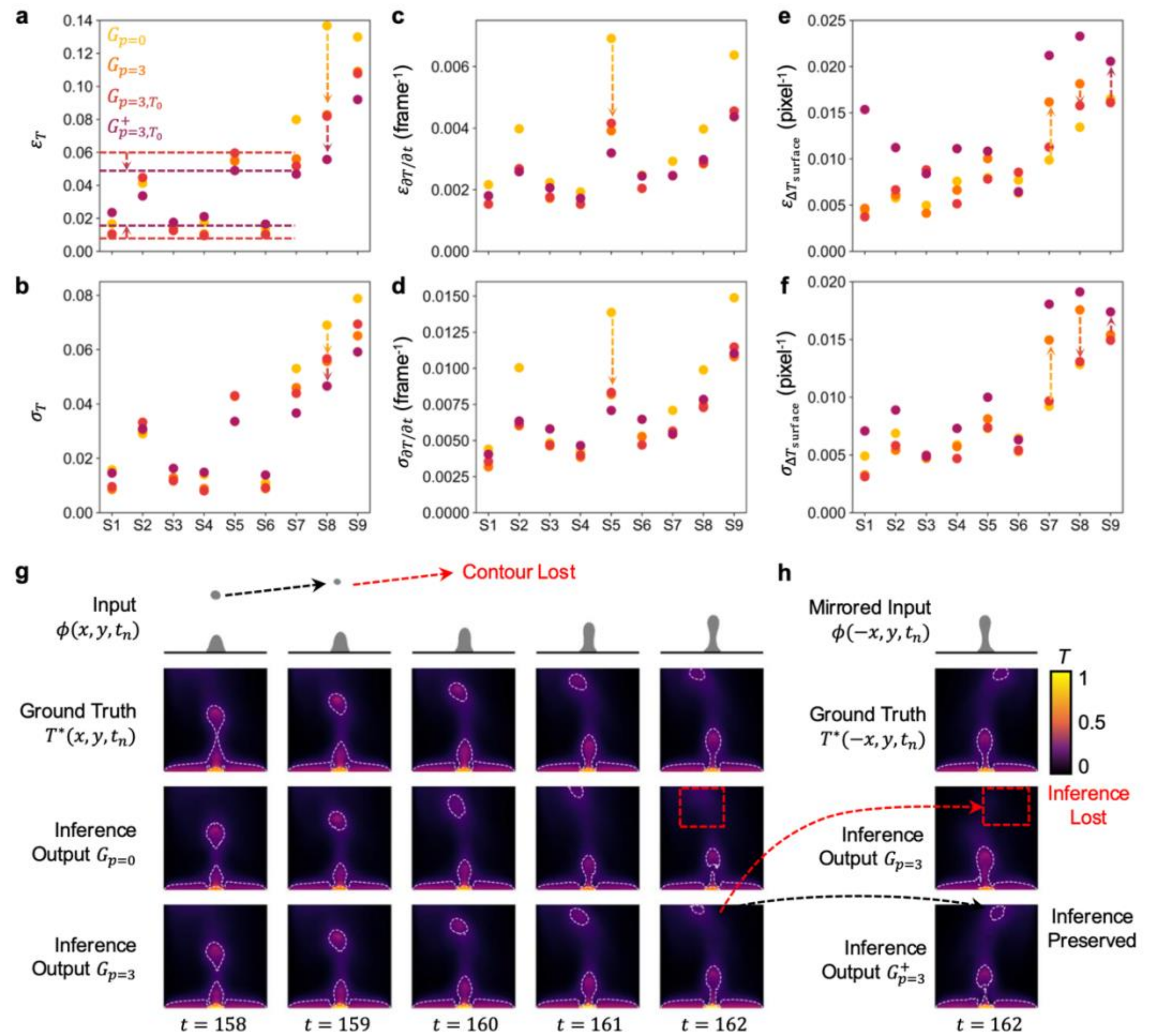

**Fig. 3.** Effect of different training schemes. The error $\varepsilon$, and the associated standard deviation $\sigma$, in (a) (b) vapor temperature, (c) (d) forward fluid temperature difference in time and (e) (f) temperature gradient at the surface is plotted for each of the 9 simulation datasets. Effect of (g) input group and (h) data augmentation on model inference accuracy for simulation datasets. Arrows are notated in (a)-(f) to specify the direction in which the error and variance shift with each additional modification to the model training. Dataset S2 is used as an example for (g) and (h). Time $t$ represents the frame number. White dashed contour lines enclose regions with a normalized temperature between 0.2 and 0.6. Red dashed boxes indicate missing inferences compared to the ground truth temperature distribution.

To increase generalizability, we further applying common data augmentation techniques such as image flipping to ensure the generators don't necessarily rely on a fixed perspective for inference. Figs. 3(a) and 3(b) show that augmentation further decreases inference error in the test datasets (S2, S5, S7-9) with slight increase in error in training datasets (S1, S3, S4, S6) as a tradeoff. Fig. 3(h) shows qualitatively that with data augmentation, the generator can preserve overall features of the temperature field even if the input image is flipped. This indicates that data augmentation is effective in reducing the directional bias in the inference result, making the generator more reliable in making physically compatible inferences. As expected, data augmentation does not show any effect on inference error temporally since it does not introduce additional temporal information. Moreover, data augmentation increases inference error in gradient, as shown in Figs. 3(g) and 3(h), since models with data augmentation cannot rely on the default direction of bubble trajectory seen in the dataset to increase inference accuracy unlike models without data augmentation. Despite this, augmented models remain relatively accurate compared to other trained generators.

We can further assess inference results of individual generator based on how well models can reproduce global trends and features observable in simulation datasets. Figs. 4(a)-(d) show two examples of such trends on different thermal variables with respect to the spatiotemporally averaged void fraction $f$, computed using the following equation:

$$f = \frac{1}{N}\sum_{t=0}^{N-1}\left(\frac{A_g}{A_l + A_g}\right)_t = \frac{1}{N}\sum_{t=0}^{N-1}\left(\frac{A_g}{A - A_s}\right)_t \tag{8}$$

where $A_g, A_l, A_s, A$ are the vapor area, liquid area, solid area, and the entire region of interest, respectively in each frame $t$ with $N$ frames total. We observe that all three variables, the excess temperature $T_{\text{excess}} = T_{\text{surface}} - T_{\text{sat}}$, the surface temperature difference (i.e., the difference between the surface temperature and the adjacent fluid), and the temperature within all three phases, increase with void fraction in our simulation datasets with varying degree. The inference result generated through both $G_{p=0}$ and $G^{+}_{p=3,T_0}$ also show qualitatively similar trends, with $G^{+}_{p=3,T_0}$ exhibits closer agreement with the ground truth trends. Furthermore, the relative order of the inferred temperature within each phase is also qualitatively respected, with solid temperature at the highest, vapor temperature at the second highest, and liquid temperature at the lowest and close to saturation temperature (i.e., 0 in normalized temperature unit) for every dataset.

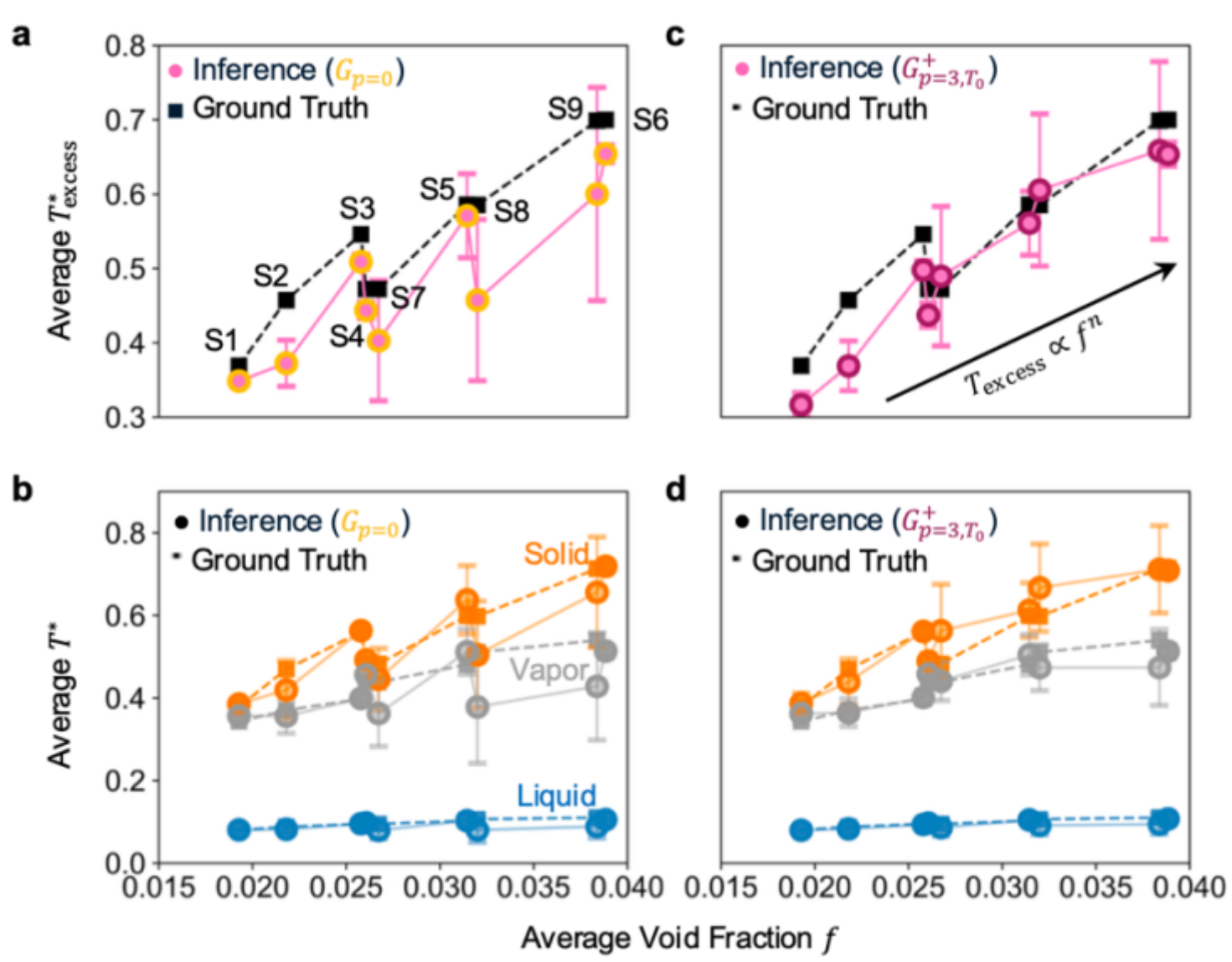

**Fig. 4.** Trends related to the void fraction observed in all 9 simulation datasets. The inferred normalized temperature are generated with $G_{p=0}$ (left column) and $G^{+}_{p=3,T_0}$ (right column). The corresponding simulation dataset numbers are listed at the top. (a) (c) The ground truth and inferred value of the spatiotemporally averaged (i.e., across the solid surface and all frames) excess temperature. (b) (d) The ground truth and inferred value of the spatiotemporally averaged (i.e., across the entire respective region and all frames) temperature in solid, vapor, and liquid phases.

### 4.2 Evaluation on Experimental Data

Since it has been known to be challenging to measure ground truth temperature profile experimentally, we can rely on the trends exhibited by the inference result to evaluate the physical plausibility of the models. Here, we recreate similar trends plotted in Figs. 4(a)-(d) in Figs. 5(b)-(g). Void fraction however is dependent on the preprocessing of the high-speed images recordings. Specifically, due to the inherent differences in the spatial domain size of the simulation and experimental data, determining a suitcase size for the input phase contour maps from experiment is crucial for accurate inference. One direct way is to match the size of the phase maps from experiment in physical unit to the size of the simulation phase maps used in training in physical unit as well (see SI-1). Here, we define the match in physical unit as applying a resizing factor of $z = 1$. Examples of the input with different $z$ are shown in Fig. 5(a).

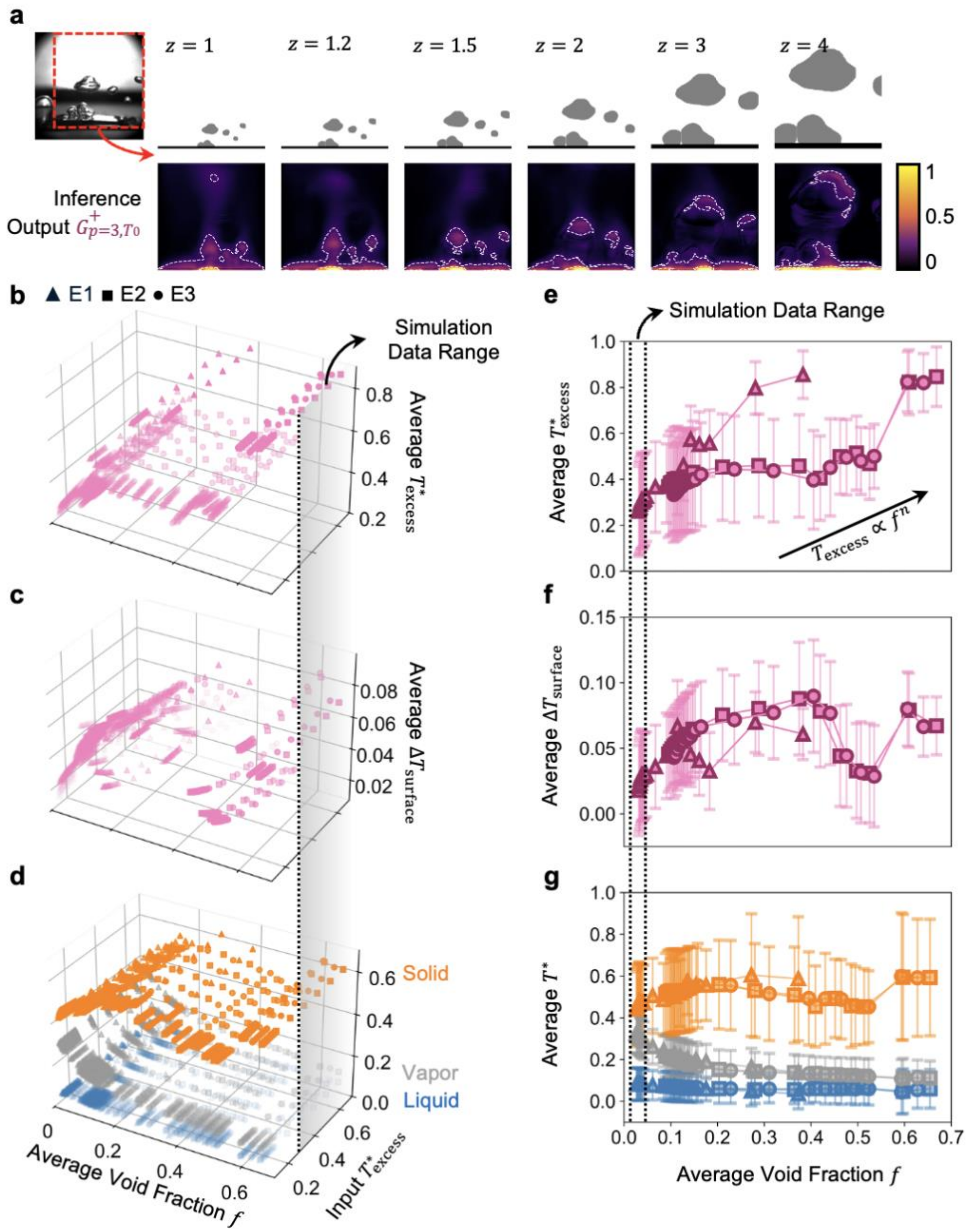

**Fig. 5.** Inferred trends by $G^+_{p=3,T_0}$ from all 3 experimental datasets. The transparency of the scattered points in (b)-(d) indicates the value (smaller = more transparent). All temperature values are in normalized unit. (a) Examples of the effect of input resizing with varying $z$. Red dashed box indicates the region with Mask R-CNN is applied and thus has meaningful segmented contours. White dashed contour lines enclose regions with a normalized temperature between 0.2 and 0.6. Trends of (b) spatiotemporally averaged excess temperature, (c) spatiotemporally averaged surface temperature difference, and (d) spatiotemporally averaged temperature within each phase are plotted with respect to both the spatiotemporally averaged void fraction $f$ (which depends on the resizing factor $z$) and the input reference temperature. Slices of the trends shown in (e)-(g) are generated by fixing the input reference temperature map according to Table 2. For (b)-(g), the void fraction $f$ of each dataset is varied by applying different zoom factors ranging from 1 to 5 (selected $z$ values: [1.0, 1.02, 1.04, 1.06, 1.08, 1.1, 1.12, 1.14, 1.16, 1.18, 1.2, 1.25, 1.5, 1.75, 2.0, 2.25, 2.5, 2.75, 3, 4, 5]). Note that applying a zoom factor of 1 requires padding artificial liquid regions around the field of view.

We can observe that although the temperature profile remains qualitatively similar with different $z$, the temperature distribution becomes more diffused (i.e., have less concentrated hotspots) as the zoom factor increases. On the other hand, the solid region doesn't diffuse as much in size, and thus the temperature distribution within solid region remains similar as $z$ varies. Similar conclusions can also be made from Figs. 5(d) and 5(g), where the temperature within the liquid and vapor region steadily converges to similar values due to diffusion as zoom factor (i.e., void fraction) increases. This diffusive effect shows that an appropriate zoom factor must be chosen for further quantitative evaluation.

For qualitative features, we can still note several features and trends that are preserved in the experimental case, for example, the ordering of the temperature values within each phase, where solid temperature is greater than vapor temperature, which is then greater than the liquid temperature in general. The excess temperature and the temperature within the solid region also exhibit an upward trend with respect to the void fraction as shown in Figs. 5(e) and 5(g).

It should be noted however that the trend is not perfectly recreated specifically at large void fraction. This is due to the drastic difference between the training (simulation) data and testing (experimental) data. Specifically, the range of void fraction covered by the training data only goes from 0.015 to 0.04, which is significantly smaller than the range covered by the testing data (Figs. 5(e)-(g)), with datasets E2 and E3 begins with an average void fraction of 0.1 even with a zoom factor of 1. This significant difference in bubble dynamics renders the learned trends unobservable, and other effects such as the diffusive effect becomes more pronounced. Similar issue can also be noted for the effect of input reference temperature. While the inferred excess temperature shows close match with the input reference temperature for simulation datasets (Fig. 4(c)), for experimental datasets there is little to no variation in inference with respect to the inference temperature (Figs. 5(b)-(d)) due to the large difference in range, with simulation data at generally high excess temperature (0.3-0.7) and experimental data at generally low excess temperature (0.1-0.15).

### 4.3 Discussion

We have proposed and developed a deep learning framework for generating temperature maps from high-speed optical images driven by simulation under similar conditions in the context of pool boiling. This framework provides a simple alternative method to generating data without involving data such as velocity maps that are difficult to collect experimentally at a high spatial and temporal resolution. We in addition have evaluated the effectiveness of several techniques in addressing unique difficulties that arise from our use case. Specifically, to minimize the loss of correlation due to the absence of phase contour due to quantization or segmentation error, we implement a many-to-one image-to-image inference scheme instead of the traditional one-to-one translation, which also greatly enhances temporally consistency. Furthermore, we employ conventional data augmentation to compensate for the lack of shared physical constraints for both simulation and experiments, which improves inference generalization (Figs. 3(a) and 3(b)). Finally, foreseeing the mismatch between simulation and experimental datasets, we test the limitation of the model by incorporating simulation and experimental data that greatly differ in thermodynamic conditions and boiling regimes. Our current implementation is still able to achieve an inference error below 0.06 (i.e., 6% of the unity in normalized unit) on average across all simulation datasets with the best selected model, and preserve several physical trends related to void fraction from the simulation datasets to the experimental datasets despite these difficulties. Errors mainly arise from persistent loss of correlation and large difference between simulation data and experimental data, such as the difference in void fraction (related to bubble behaviors and boiling regimes) among other factors including fluid properties (related to saturation temperature), heater arrangement, spatial and temporal resolution, and initial and boundary conditions that are not explicitly available during training time.

# 5. Conclusion

We have shown that even with limited training data, no strict enforcement of physics and no dedicated model optimization, generative deep learning method is still capable of producing qualitatively accurate inferences on relevant thermodynamic variables such as temperature distribution within each phase region. Our implementation however revealed the limitation of such approach that stems from the lack of diversity of the training datasets. Figs. 6(a)-(e) highlight the significant variation in the geometries of the phase contour across the nucleate boiling regime. From our training result, we observe that inference error is relatively large when there is a significant change in the bubble behavior, as shown in Figs. 6(a) and 6(b). Change in bubble behavior also drives the inference error on experimental datasets, since the bubble behavior observed in experimental dataset shows continued bubble nucleation and departure instead of isolated bubble nucleation on the surface seen in simulation datasets, as shown in Figs. 6(c) and 6(d). Moreover, we show that the inference results can be significantly influenced by simply resizing the geometry. Since with large resizing factor, the bubble contour is similar to those observed in pool boiling at high heat fluxes (Fig. 6(d)), this implies that our model will produce invalid inferences for pool boiling at high heat fluxes, which is beyond the regimes exhibited by the training datasets. Although model performance is expected to improve by simply incorporating additional training datasets with different boiling behavior, acquiring simulated and experimental datasets under different boiling conditions remains a fundamental bottleneck for our deep learning inference framework.

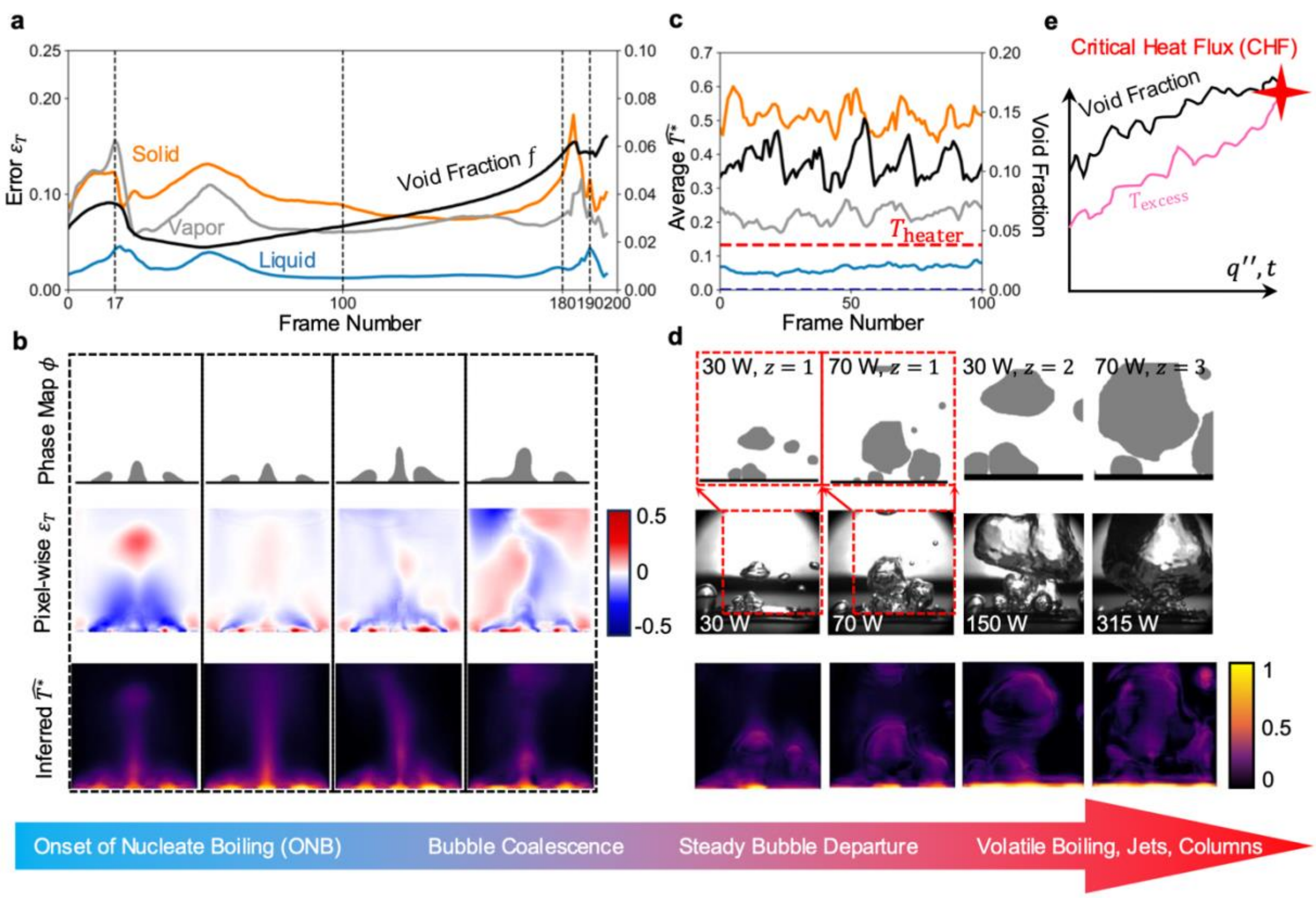

**Fig. 6.** Impact of the diversity of the pool boiling datasets on model inference results. Model $G^+_{p=3,T_0}$ is used for all inference results shown. (a) Inference error on simulation dataset S8 over time. (b) Phase contours and the associated inference error map and inferred temperature maps at key frames ($t = 17, 100, 180, 190$) labelled in (a). (c) Inferred temperature variation over time on experimental dataset E1 with $z = 2$ and a reference temperature map constructed using the condition in Table 2. (d) Examples of the experimental dataset and the associated inference

results. For higher heat fluxes, example phase contours with larger zoom factor are chosen for comparison. (e) Overall projected trend of the variation in void fraction and excess temperature across the remaining nucleate boiling regime.

## Acknowledgements

Q.F. is thankful for the financial support from the UC Irvine Mechanical and Aerospace Engineering Department for graduate student researcher. The authors gratefully acknowledge funding support from the Office of Naval Research (ONR) MURI under Grant No. N000142412575 with Dr. Mark Spector serving as the program officer, and funding support from the National Science Foundation (NSF) under Grant No. 2045322.

## CRediT Authorship Contribution Statement

**Qianxi Fu**: Methodology, Software, Formal Analysis, Writing – Original Draft, Writing – Review & Editing, Visualization. **Youngjoon Suh**: Conceptualization, Methodology, Software, Investigation, Writing – Review & Editing, Supervision. **Xiaojing Zhang**: Investigation, Data curation. **Sanghyeon Chang:** Writing – Review & Editing, **Yoonjin Won**: Conceptualization, Writing – Review & Editing, Supervision, Project Administration, Funding Acquisition.

## Conflict of Interests

There are no conflicts to declare.

## Nomenclature

| | | | |
|---|---|---|---|
| $\boldsymbol{A}$ | Total 2D view area (mm$^2$ or lu$^2$) | Greek Letters | |
| $\boldsymbol{A_g}$ | Gas (vapor) area (mm$^2$ or lu$^2$) | $\phi(x, y, t)$ | Phase contour |
| $\boldsymbol{A_l}$ | Liquid area (mm$^2$ or lu$^2$) | $\Phi(x, y, t)$ | Input layers/stack |
| $\boldsymbol{A_s}$ | Solid area (mm$^2$ or lu$^2$) | $\lambda$ | Weight assigned to loss components |
| $\boldsymbol{f}$ | Void fraction | $\kappa$ | Thermal conductivity (W/m·K) |
| $\mathbf{Ja}$ | Jakob Number | $\varepsilon$ | Error |
| $\boldsymbol{N}$ | Final frame number | $\sigma$ | Standard deviation |
| $\boldsymbol{p}$ | Number of previous frames | | |
| $\boldsymbol{q''(x, y_h, t)}$ | Local surface heat flux (W/cm$^2$) | | |
| $\boldsymbol{q''_S(t)}$ | Spatially averaged surface heat flux (W/cm$^2$) | | |
| $\boldsymbol{q''_{ST}(t)}$ | Spatiotemporally averaged surface heat flux (W/cm$^2$) | | |
| $\boldsymbol{T}$ | Temperature (K or tu) | | |
| $\boldsymbol{T^*}$ | Normalized Jakob Number | | |
| $\boldsymbol{T_c}$ | Critical temperature (K or tu) | | |
| $\boldsymbol{T_{\mathrm{sat}}}$ | Saturation temperature (K or tu) | | |
| $\boldsymbol{T_{\mathrm{ref}}}$ | Reference temperature (K or tu) | | |
| $\boldsymbol{x, y}$ | Coordinates (mm or lu) | | |
| $\boldsymbol{y_h}$ | Vertical heater surface coordinate (mm or lu) | | |

# Supplemental Information

## 1. Simulation Data Generation, Characterization & Processing

**1.1 Simulation Methods**
We use a hybrid scheme with pseudopotential LBM with single relaxation time (SRT) and finite difference method (FDM), similar to the setup in [1], to generate pool boiling simulation data used in this paper.

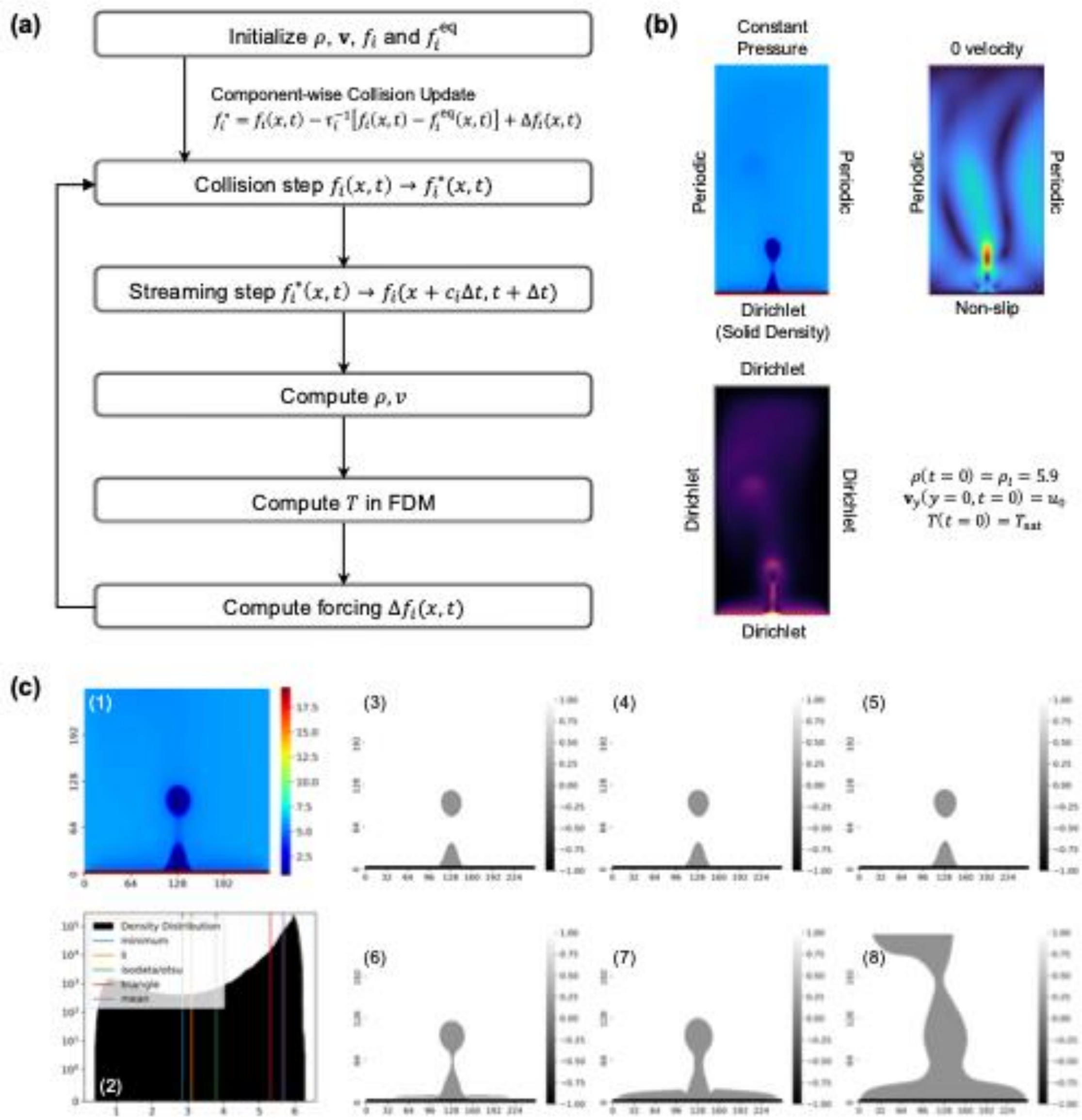

**Fig. S1**: LB-FD simulation data details. (a) LB-FD simulation flowchart. (b) Imposed initial and boundary conditions. The (upward) initial velocity is only applied at the solid surface, with everywhere else at 0 velocity. (c) Effects of different quantization methods. (1) original density map, (2) probability density distribution of the density values in the training datasets (dataset 1,3,4,6), (3) $\rho_{\text{th}} = 2.85$ (minimum), (4) $\rho_{\text{th}} = 3.11$ (Li), (5) $\rho_{\text{th}} = 3.80$ (ISODATA, Otsu), (6) $\rho_{\text{th}} = 4.90$, (7) $\rho_{\text{th}} = 5.31$ (triangle), (8) $\rho_{\text{th}} = 5.685$ (mean).

We conduct 2D pool boiling simulation according to the flowchart shown in Fig. S1(a) using two-dimensional lattice with nine unit lattice vectors (D2Q9). Each simulation frame has a resolution of 512 x 256 lattice units (lu). The initial and boundary conditions imposed on each variable are shown in Fig. S1(b). Note that for our objective, simulation data is cropped to 256 x 256 during pre-processing for the purpose of squaring the input and creating a free boundary at the top, reducing processing load and matching partially with the experimental data.

During simulation, the probability distribution function $f_i(x, v, t)$ along each lattice direction $i$ is updated due to collision and streaming, from which the density $\rho$ and velocity fields $\mathbf{v}$ are computed. Thermal simulation is coupled with LBM through the following energy equation:

$$\frac{\partial T}{\partial t} = -\mathbf{v} \cdot \nabla T + \frac{1}{\rho c_v} \nabla \cdot (\kappa \nabla T) - \frac{T}{\rho c_v} \left( \frac{\partial P}{\partial T} \right)_\rho \nabla \cdot \mathbf{v}$$

where $c_v$ is the heat capacity at constant volume of the liquid and $\kappa$ is the thermal conductivity. The energy equation is advanced through FDM with second-order and fourth-order Runge-Kutta method applied to spatial and temporal discretization respectively. The pressure term $P$ is related to density and temperature through the Peng-Robinson (P-R) equation of state:

$$P = \frac{\rho R T}{1 - b\rho} - \frac{a \rho^2 \varepsilon(T)}{1 + 2b\rho - b^2 \rho^2}, \qquad \begin{cases} a = 0.45724 R^2 T_c^2 / P_c \\ b = 0.0778 R T_c / P_c \end{cases}$$

where $P_c$ and $T_c$ denote the critical pressure and temperature respectively. The ideal gas constant $R$ is taken to be 1 in LB unit, and $a = 2/49$ and $b = 2/21$ are used. The expression $\varepsilon(T)$ is given by:

$$\varepsilon(T) = [1 + (0.37464 + 1.54226\omega - 0.26992\omega^2)(1 - \sqrt{T/T_c})]^2$$

Here $\omega = 0.344$ is taken, since it has been shown that with this value the coexistence curve derived from the above equation of state matches closely with the experimental data for saturated water-steam system [?]. Finally, the forcing term is updated based on pseudopotential formulation:

$$F_{int}(x) = -\Psi(x) \sum_{x'} G(x, x') \Psi(x')(x' - x), \qquad \Psi(x) = \sqrt{\frac{2(P - \rho c_s^2)}{c_0 g}}$$

All relevant constants used in the LBM simulation are listed in Table S1 in terms of both the LB unit and the corresponding physical unit. Unit conversion is based on the principle of corresponding states [2], where any value $x$ in physical unit with its respective critical value $x_c$ (or characteristic value with a zero subscript) follows this relationship:

$$\frac{x}{x_c} = \frac{x_{LB}}{x_{c,LB}}$$

where the $LB$ subscript denotes values in LB units. Certain values such as heat capacity and latent heat of vaporization do not have a critical value for conversion. Their values in physical units are directly referenced from tables at saturation temperature when the calculation requires.

**1.2 Determination of Quantization Threshold**

The phase contour maps are generated through density map quantization or thresholding. We derive a global threshold value $\rho_{th}$ by applying built-in methods such as triangle and ISODATA thresholding in python to training datasets (i.e. simulation dataset 1,3,4 and 6) only. Fig. S1(c) shows the effect of different threshold values on density maps. For the results presented in the main text, we have selected the threshold of $\rho_{th} = 3.79552$ (from ISODATA or Otsu's methods) simply based on a qualitative assessment of the resulting phase contour maps (i.e. the bubble outline follows our intuitive assignment) rather than its potential impact on model training and inference.

**Table S1**
Relevant constants used in simulation in both LBM and physical units. The LBM lattice unit for mass, length, time and temperature are mu, lu, ts, and tu, respectively.

| Simulation Constants | LBM Unit | Physical Unit |
|---|---|---|
| $\boldsymbol{T_c}$ | 0.0729 tu | 647.17 K |
| $\boldsymbol{\rho_c}$ | 2.657304 mu/lu$^3$ | 322 kg/m$^3$ |
| $\boldsymbol{u_0}$ | 0.02627 lu/ts | 0.1645 m/s |
| $\boldsymbol{l_0}$ | 23.00421 lu | 2.762×10$^{-3}$ m |
| $\boldsymbol{t_0}$ | 875.6753 ts | 0.01679 s |
| $\boldsymbol{g}$ | 3×10$^{-5}$ lu/ts$^2$ | 9.81 m/s$^2$ |
| $\boldsymbol{\sigma}$ | 0.09716 | 74.8×10$^{-3}$ N/m |

## 2. Experimental Data Pre-Processing

We follow the flowchart shown in Fig. S2 to extract phase contour maps from high-speed images. We first use a fine-tuned Mask R-CNN model to perform instance segmentation on high-speed recordings to isolate vapor region [3]. The Mask R-CNN model was trained specifically for pool boiling using a meticulously curated dataset of 5,555 training images and 1,388 test images. These images were collected over several years from diverse experimental setups to capture the wide-ranging complexities of pool boiling phenomena. A team of trained annotators followed detailed project-specific guidelines to ensure consistent and accurate labeling. Regular communication among annotators helped resolve ambiguities and refine annotation standards, further enhancing the quality of the dataset. The annotation process was managed using Supervisely (San Jose, CA, USA), a commercial platform designed for efficient labeling workflows and employed human-in-the-loop techniques to iteratively address edge cases and challenging scenarios. To improve the model's generalizability, the dataset was diversified by incorporating images captured under various experimental conditions, such as different heat flux levels and surface properties. Additional variability was introduced using data augmentation techniques, including horizontal flips, brightness adjustments, contrast modifications, and resizing. These augmentations effectively expanded the dataset, simulating a broader range of conditions and improving the robustness of the trained model. The Mask R-CNN model was trained using stochastic gradient descent with a learning rate of 0.0008, a momentum of 0.9, and over 100 epochs. These training parameters were optimized to balance convergence and stability, resulting in a model that achieved a test loss of 0.046, demonstrating high accuracy in bubble detection and segmentation.

The outputs of the trained Mask R-CNN model are 8-bit instance-specific masks, with each bubble instance assigned a unique intensity value ranging from 1 to 256, representing its unique ID. Image masks then are cropped to heater (bubble-surface contact line) and only the bubble masks generated from the heater are kept. Images are then cropped and resized to a resolution of 256 x 256 with the same quantized values assigned to each phase region. The required dimension $X$ such that the size of each pixel in the experimental datasets match that in the simulation datasets in physical units after resizing is:

$$X = \frac{256(\Delta x_{LB})}{\Delta x} = \frac{256\left(\frac{l_0}{l_{0,LB}}\right)}{\left(\frac{L_{\text{heater}}}{L_{\text{heater},PX}}\right)}$$

where $l_0$ is the characteristic length of the simulation (Table S1), and $L_{\text{heater}} = 10$ mm is the length of the heater with an original pixel length of $L_{\text{heater},PX} = 520$ pixels. In our case, matching the pixel size may require a resolution $X$ greater than 1024. We instead pad the image to 2048 x 2048 with liquid cells at saturation temperature after cropping so that sufficient pixels are available for resizing. The solid layer is added after resizing with a thickness of 5 pixels and spanning the full length of 256 pixels.

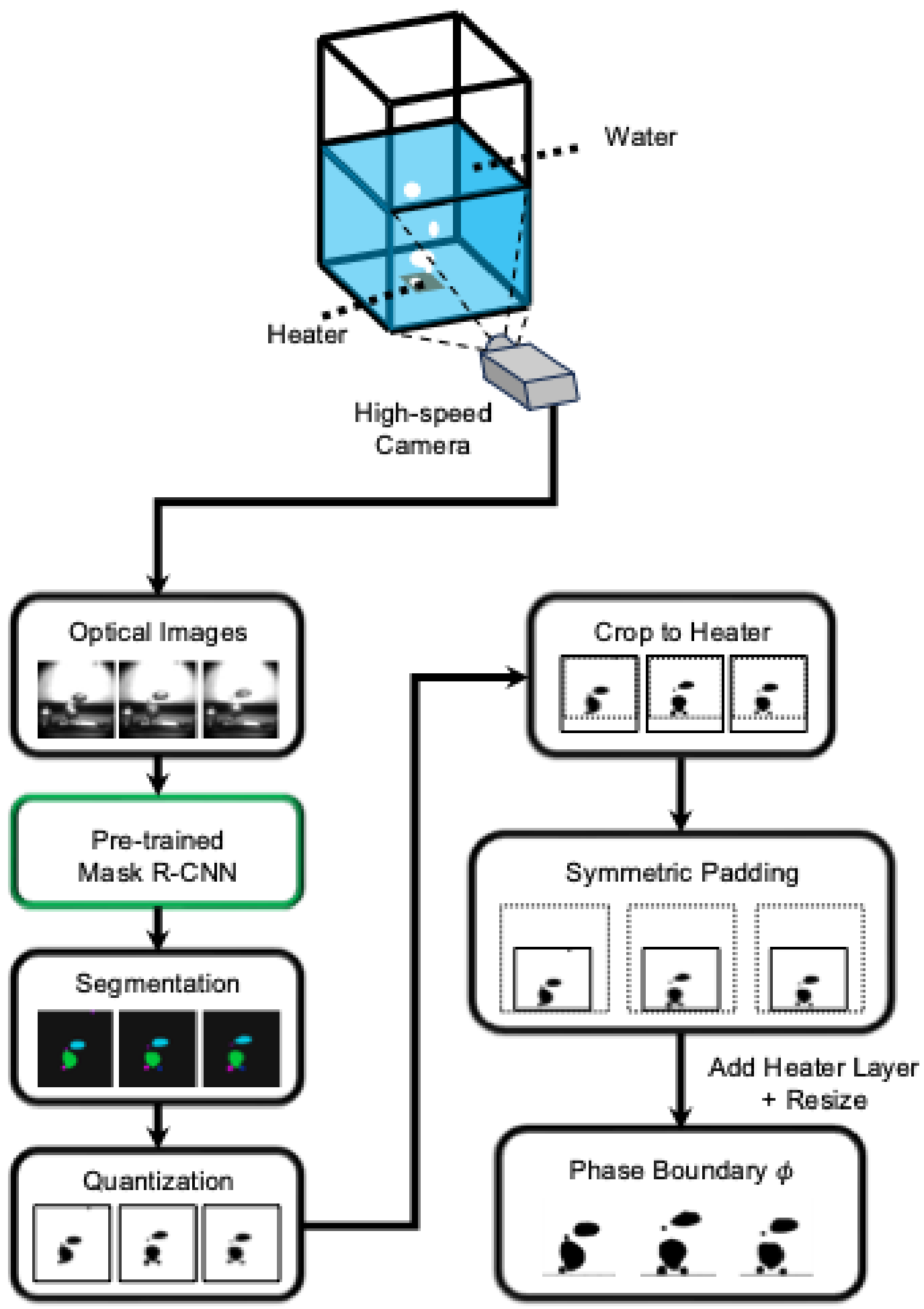

**Fig. S2**: Experimental data pre-processing steps. (a) Phase contour extraction. (b) Thermocouple arrangement and heating variations.

# 3. Model Training Details

### 3.1 Conditional GAN Structure & Selected Parameters

Figs. S3(a) and S3(b) show the structure of the generator and the discriminator, respectively. Each 2D convolution layer in the generator has a kernel size of 5 x 5, with batch normalization applied, leaky ReLU activation (except the last layer with a hyperbolic tangent activation), paddings to preserve dimension and no bias term. The 2D convolution layers in the discriminator follow similar implementation, except the kernel sizes are 7 x 7, 5 x 5, 3 x 3, 5 x 5 for each layer, respectively. The initializer in both the generator and the discriminator is a random initializer sampled from a normal distribution with mean of 0 and a standard deviation of 0.01, except the last layer in both which uses a He normal initializer. We use up-sampling layer followed by a stride-1 2D convolution to avoid checkerboard artifacts that result from 2D transposed convolution [4].

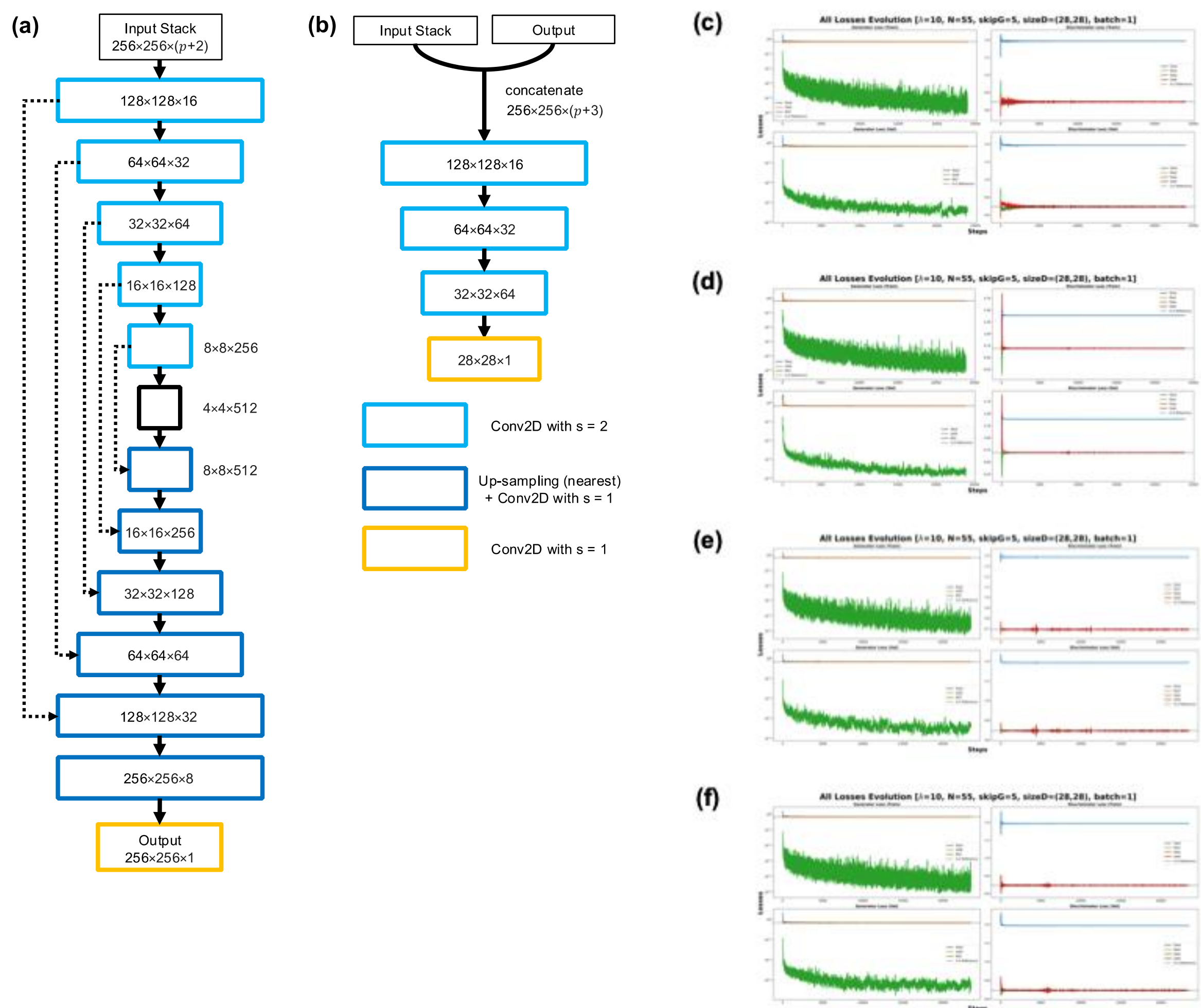

**Fig. S3**: Model training details. (a) Generator structure. (b) Discriminator structure. Training losses evolution of (c) model $G_{p=0}$, (d) model $G_{p=3}$, (e) model $G_{p=3,T_{\text{ref}}}$, (f) model $G^{+}_{p=3,T_{\text{ref}}}$.

### S3.2 Training Losses

Each model is trained with a batch size of 1. We use Adam optimizer [5] with a learning rate of 0.0001 and default momentum parameters of $(\beta_1, \beta_2) = (0.9, 0.999)$. The training losses over 30 epochs (23610 steps) for each model are shown in Figs. S3(c-f). The adversarial losses should fluctuate about ln 2 to ensure continuous learning. During training, we find the discriminator usually learns faster than the generator, as a result we train the generator twice at each step instead of one.

# 4. Additional Results

### 4.1 Inference Results in Video Format

Two videos showing the inference results are attached separately. Video “Inference S8” uses dataset S8 as input, and the resulting inferred temperature as well as inference error and surface temperature gradient (both ground truth and the inferred) are shown in the video. Video “Inference E1” uses dataset E1 as input. Both videos are generated with Model $G^{+}_{p=3,T_{\mathrm{ref}}}$.

### 4.2 Model Sensitivity to Input Conditions

Figs. S4, S5 and S6 show the inference heatmap on dataset E2 by $G^{+}_{p=3,T_{\mathrm{ref}}}$ which highlights the effect of variation in input reference temperature and zoom factor on inference results. Variation is more significant at large zoom factors.

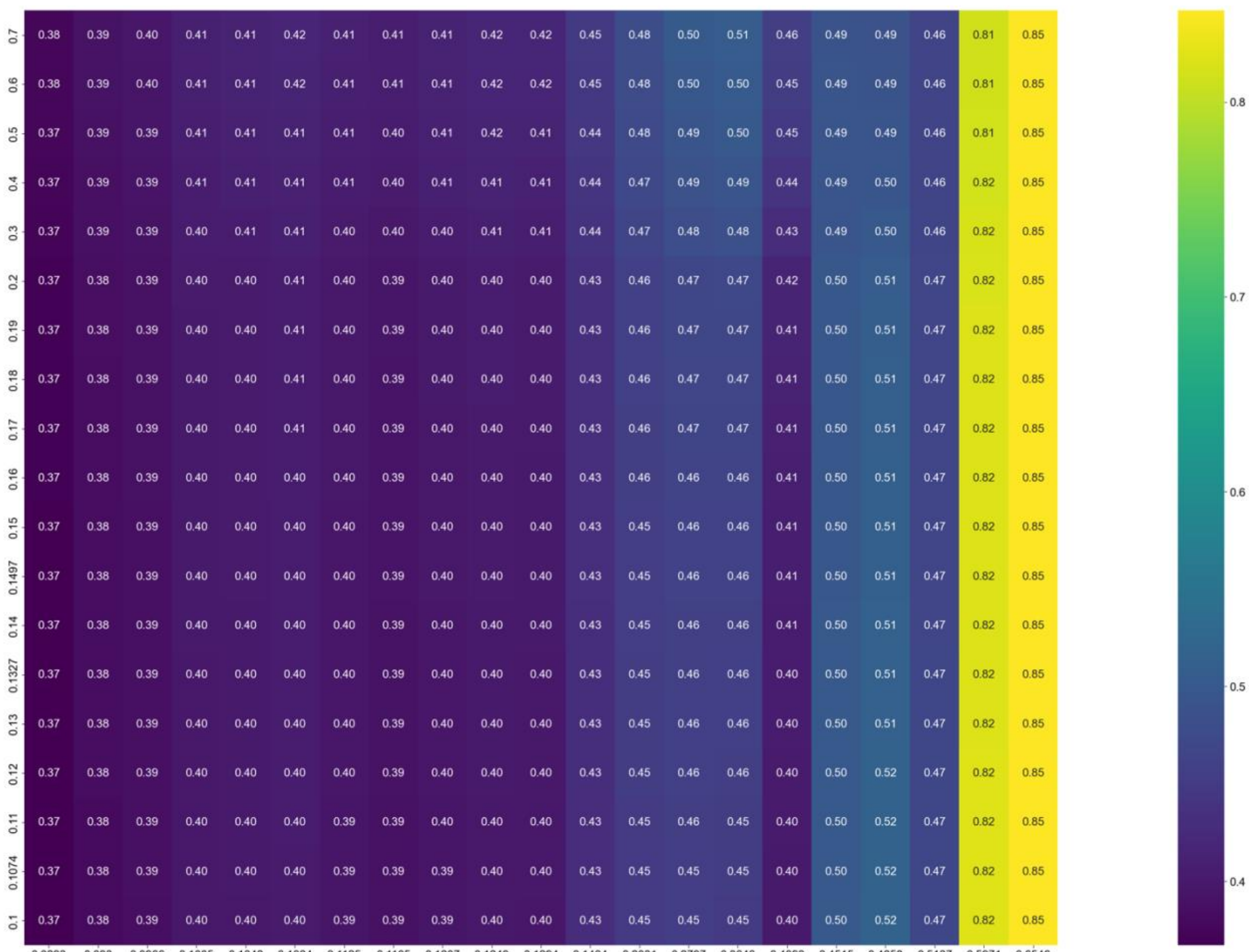

**Fig. S4**: Heatmap of the average inferred solid temperature with dataset E2. Horizontal axis represents varying average void fraction $f$ due to change in zoom factor ($z$ = [1.0, 1.02, 1.04, 1.06, 1.08, 1.1, 1.12, 1.14, 1.16, 1.18, 1.2, 1.25, 1.5, 1.75, 2.0, 2.25, 2.5, 2.75, 3, 4, 5]) and vertical axis represents varying $T^{*}_{\mathrm{ref}}$ ranging from 0.07 to 0.7.

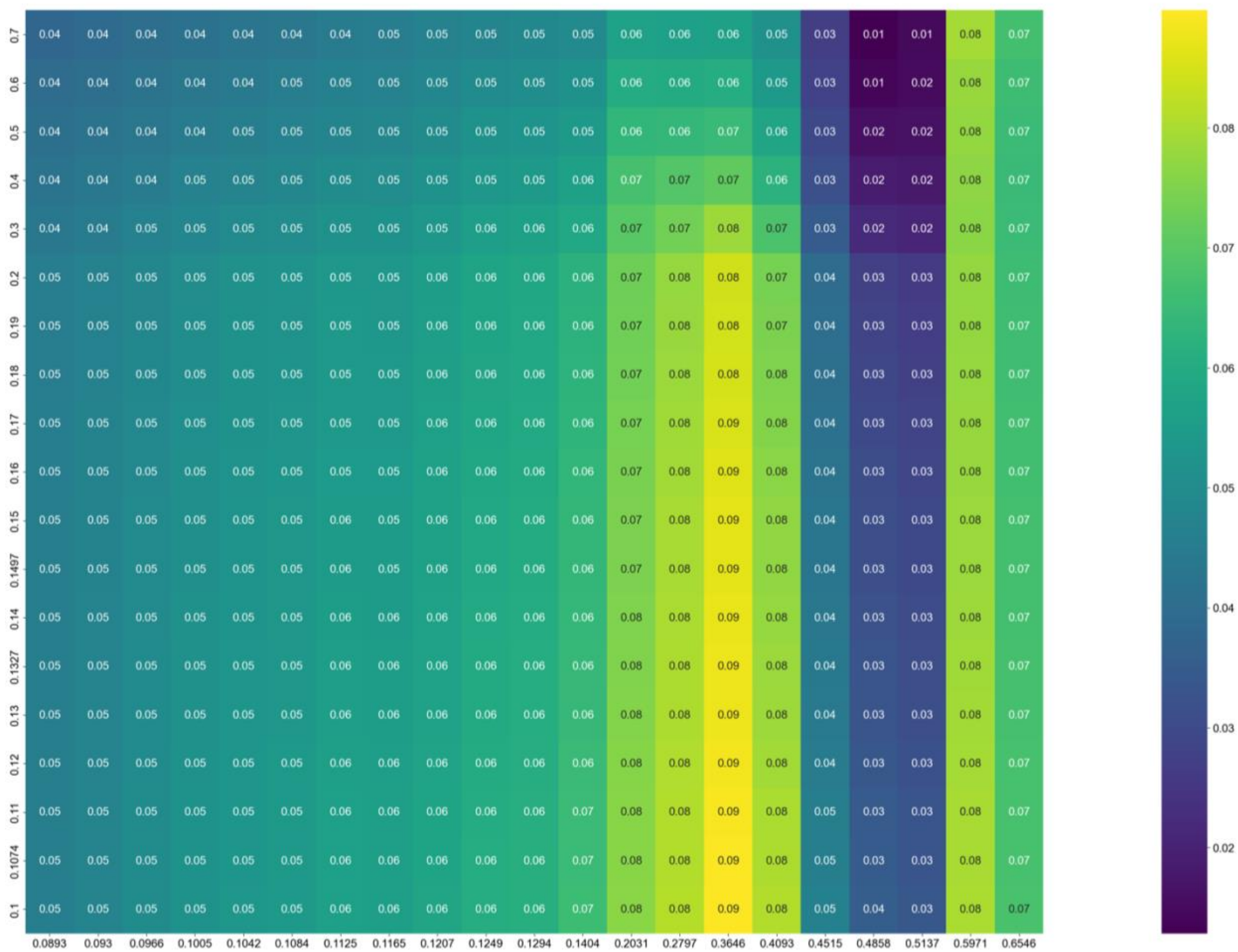

**Fig. S5**: Heatmap of the average inferred liquid temperature with dataset E2. Horizontal axis represents varying average void fraction $f$ due to change in zoom factor ($z$ = [1.0, 1.02, 1.04, 1.06, 1.08, 1.1, 1.12, 1.14, 1.16, 1.18, 1.2, 1.25, 1.5, 1.75, 2.0, 2.25, 2.5, 2.75, 3, 4, 5]) and vertical axis represents varying $T^*_{\text{ref}}$ ranging from 0.07 to 0.7.

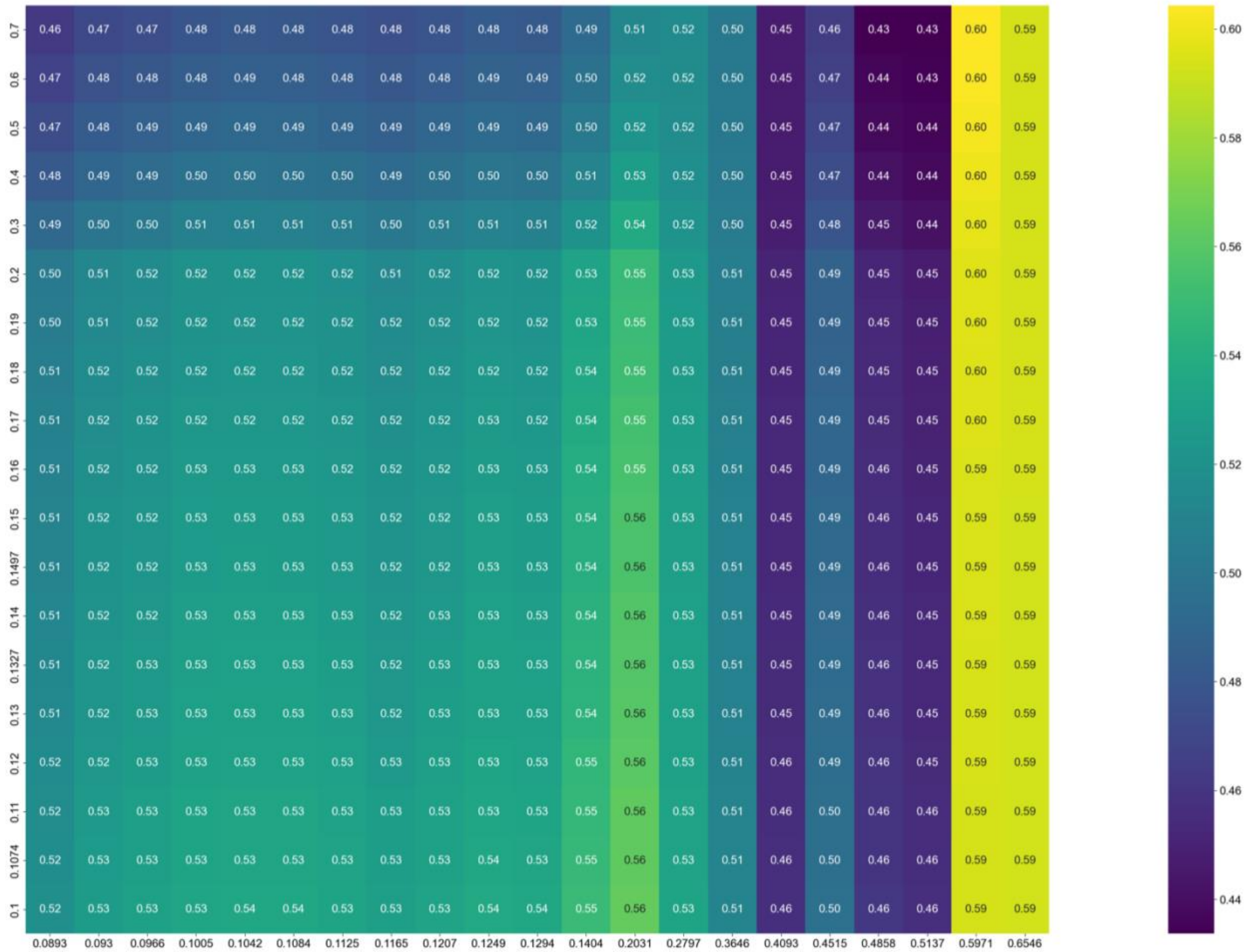

**Fig. S6**: Heatmap of the average inferred vapor temperature with dataset E2. Horizontal axis represents varying average void fraction $f$ due to change in zoom factor ($z$ = [1.0, 1.02, 1.04, 1.06, 1.08, 1.1, 1.12, 1.14, 1.16, 1.18, 1.2, 1.25, 1.5, 1.75, 2.0, 2.25, 2.5, 2.75, 3, 4, 5]) and vertical axis represents varying $T^*_{\mathrm{ref}}$ ranging from 0.07 to 0.7.

## References (Supplemental)